# A Magnetic Millirobot Walks on Slippery Biological Surfaces for Targeted Cargo Delivery

Moonkwang Jeong[1], Xiangzhou Tan[1,2], Felix Fischer[1,‡] and Tian Qiu[1,‡,*]

[1]Cyber Valley group - Biomedical Microsystems, Institute of Physical Chemistry, University of Stuttgart, Pfaffenwaldring 55, 70569 Stuttgart, Germany

[2]Department of General Surgery, Xiangya Hospital, Central South University, Changsha 410008, China; xiangzhou.tan@csu.edu.cn

[‡]Current affiliation: Department of Smart Technologies for Tumor Therapy, German Cancer Research Center Site Dresden, Blasewitzer Str. 80, 01307 Dresden, Germany

*Correspondence: tian.qiu@dkfz-heidelberg.de

**Abstract**: Small-scale robots hold great potential for targeted cargo delivery in minimally-invasive medicine. However, current robots often face challenges to locomote efficiently on slippery biological tissue surfaces, especially when loaded with heavy cargos. Here, we report a magnetic millirobot that can walk on rough and slippery biological tissues by anchoring itself on the soft tissue surface alternately with two feet and reciprocally rotating the body to move forward. We experimentally studied the locomotion, validated it with numerical simulations and optimized the actuation parameters to fit various terrains and loading conditions. Furthermore, we developed a permanent magnet set-up to enable wireless actuation within a human-scale volume which allows precise control of the millirobot to follow complex trajectories, climb vertical walls, and carry cargo up to four times of its own weight. Upon reaching the target location, it performs a deployment sequence to release the liquid drug into tissues. The robust gait of our millirobot on rough biological terrains, combined with its heavy load capacity, make it a versatile and effective miniaturized vehicle for targeted cargo delivery.

1. Introduction

Wirelessly actuated small-scale robots show great potential for minimally-invasive medical procedures [1–3]. Besides their tiny footprints, small-scale robots exhibit high dexterity and precise controllability, as they are actuated by physical or chemical fields in a wireless way. Various physical fields have been reported to power robots, for example, magnetic fields [4–10], ultrasound fields [11–13] and light fields [14]. Among all actuation schemes, magnetism is one of the most common modalities used for biomedical applications due to its low attenuation in biological tissues, strong actuation force and precise controllability. For example, helical micropropellers can swim in biological fluids [4–6], soft robots use multimodal locomotion for cargo transportation [7,8], and biped robot can walk on a substrate surface [9]. A capsule robot at centimeter scale was demonstrated to controllably move on the surface of a porcine stomach for endoscopy and also perform a needle biopsy [10]. Despite these advances, there are still many major challenges for small-scale robots to be applied in real medical scenarios in the human body.

First, real biological environments pose a challenge for a small-scale robot to locomote as they are mostly viscoelastic and difficult to penetrate. The tissue surfaces often exhibit complicated three-dimensional (3D) shapes and features, and are coated with slippery biological fluids. Most current small-scale robots are designed and tested under idealized lab conditions, such as propulsion in water and crawling on flat and dry surfaces. Hence, these results often do not readily transfer to useful schemes in real medical applications. Additionally, they usually lack sufficient grip to the slippery biological surface, which inhibits efficient crawling of the robot. Sharp tips that penetrate the superficial layer of tissues have been used in medical implants and larger scale crawling robots to anchor to soft tissues [15,16], however, the anchorin g mechanism has not been widely considered for millirobots due to its difficulties in micro-fabrication and characterization. Here, we tackle this challenge by fabricating 3D sharp metallic tips as robot's feet for efficient anchoring to slippery biological surfaces.

Second, it is difficult to generate a sufficient magnetic field strength or gradient to actuate a robot while maintaining a large enough space to accommodate a human as the magnetic field decays significantly over distance [17]. Mainly two types of magnetic actuation systems have been used for the actuation of small robots. On the one hand, electromagnetic systems exhibit high controllability and can be completely turned off, however, they have limitations in the working volume and field strength, and also require huge amounts of electric power and cooling system. In contrast, permanent magnet systems offer a higher magnetic field strength and a bigger working volume without cooling issue, but are limited in the driving frequency and suffer from mechanical noise and vibration [18]. Here, we develop a permanent magnet set-up to wirelessly actuate the walking robot, which can incorporate a human body aiming at real medical applications in the future.

In this paper, we report a millirobot that can walk on a slippery biological surface and inject drugs into the tissue at a target location. The millirobot is actuated wirelessly by an oscillating magnetic field, generated by a set of permanent magnets, comprising a static pair of magnets and a rotating magnet. The robot has specially designed feet, which can alternatingly anchor themselves to soft biological tissues. By alternating 3D rotations around the two front feet, the robot exhibits a bipedal gait. The locomotion was experimentally measured, geometrically validated by numerical simulations, and an optimization of the walking speed on rough and slippery terrain was carried out. The robot is capable of climbing on vertical walls, carrying four times of its own weight. Changing the actuation mode of the permanent magnet, the robot is capable of performing a second motion sequence, which deploys the carried cargo, i.e., to inject liquid drug into the soft tissues. The motion of the millirobot is controlled to follow arbitrarily-designed trajectories on a 3D surface. The reported method brings new insights for designing small-scale robots crawling on rough and slippery biological surfaces and may lead to many potential applications in minimally-invasive medical procedures.

2. Methods

2.1. Design and fabrication of the millirobot

As shown in Fig. 1a, the millirobot consists of three parts: a chassis with four feet, a magnetic actuator, and a cargo, for example a capillary loaded with drugs. The chassis is made of Molybdenum (>99% purity, 50 μm thick) with four feet, i.e. sharp tips (~10 μm at the apex), fabricated by femtosecond laser cutting (MPS FLEXIBLE, ROFIN-BAASEL Lasertech, Germany) and folded towards the bottom by pressing against a metallic rod (OD 1.2 mm). The actuator is a cylindrical permanent magnet magnetized along the longitudinal direction (NdFeB-N45, OD 1 mm, Height 1 mm, Supermagnete, Germany). It was assembled on the chassis using an adhesive (Loctite 401, Henkel, Germany). The cargo was assembled to the actuator. For instance, a capillary (OD 1 mm, ID 0.8 mm, CM Scientific Ryefield Ltd.) was diced to a sharp tip by a diamond cutter and affixed to the upper edge of the robot, positioned to pierce the tissue surface when the robot is tilted over 90 degrees. This enables the robot to inject liquid drugs effectively into soft tissues at the target location.

2.2. The permanent magnetic actuation system

Fig. 1b and 1c show the magnetic actuation system. The system comprises three permanent magnets, which generate a superimposed oscillating magnetic field in the working volume at the center of the set-up. A static magnetic field $B_h$ is generated by a pair of magnets M1 and M2 (NdFeB-N45, 110.6 × 89 × 19.5 mm3, Supermagnete). And an oscillating magnetic field

Br is generated by a third magnet M3 (NdFeB-N40, 50.8 × 50.8 × 50.8 mm3, Supermagnete). The system has in total two rotational degrees of freedom as control inputs, labeled as α and β respectively. It is possible to achieve controlled rotation of M3 around the X-axis at an angle α using a stepper motor (23HS30-2804Sm, OMC Co. Ltd., China). The robot's walking direction can be controlled by rotating the magnetic actuation system around the Y-axis at a controlled angle β. The resulting magnetic field of the system, represented by the vector **B**, rotates on the lateral surface of a virtual cone as shown in Fig. 1d. The vector **B** is determined by the distances of the permanent magnets W and D (Fig. 1c), and the angle α, which defines the two rotational degrees of freedom θ and φ of the millirobot.

The robot aligns its magnetic moment to the applied external magnetic field B, thus the two angles mentioned above determine the gait of the millirobot. The actuation system was designed to exhibit a large accessible space that is suitable for future medical applications on a human patient. As shown in Fig. 1c, W (the working width between M1 and M2) and D (the distance from the surface of M3 to the center of the working volume) are larger than 50 cm and 15 cm, respectively.

2.3. Numerical simulation of the magnetic actuation system

3D numerical simulations of the magnetic actuation system were performed using the AC/DC Module in COMSOL Multiphysics (COMSOL Multiphysics, Burlington, MA, USA). The center-to-center distance from M1 to M2 and to M3 were defined as 484 mm and 136 mm, respectively. To simplify the model, the surrounding air (1000×1000×1000 $mm^3$) was defined as a magnetic insulation boundary condition. The simulation was conducted as steady-state magnetic field at the angle α from -75° to 75° with a step size of 1° for the validation of a magnetic flux density in the permanent magnetic actuation system. An integrated physics-controlled mesh was used for extremely fine element size.

2.4. Motion sequences of the millirobot: locomotion and deployment

The robot exhibits a walking motion, as the **B** field rotates on the virtual cone surface (Fig. 1d). The gait is illustrated in Fig. 2a. At the beginning, α = 0°, the robot stands at the angle of θ (see Fig. 1d). The zero degree of α is defined when the north pole of M3 points upwards along with Y-axis. When α increases up to 72°, the robot turns to the right (orange arrow) while the front right (FR) foot remains anchored to the biological soft tissue (orange circle). Reversely, when the α decreases from 72° to 0°, the robot turns to the left (green arrow) while the front left (FL) foot anchors to the tissue (green circle). By repeating the motions, the robot moves forward along the X-axis direction.

The second motion of the scheme is cargo deployment. As illustrated in Fig. 2b, when the magnetic field rotates to 180° or more, the robot flips by following the external magnetic field B. The tip of the capillary attached on the robot touches the tissue surface when α is approximately 90°. As the magnetic field keeps rotating, the robot stands upside down on the apex of the capillary. The magnetic field gradient pulls the robot towards the tissue surface, and the tip can penetrate soft tissue and inject drug into it. The robot rotates its body back to the walking motion (α = 72°) when the drug delivery is done, and can move by the walking motion to the next target location. During the motion of deployment, the robot loses its anchoring of the feet and shortly after, the capillary tip enters the tissue to deploy the cargo, which also acts as a new anchoring point to fix the robot. Hence, slipping can be avoided in both motions 1 and 2.

2.5. Testing of the millirobot on biological surfaces

The millirobot was tested on a hydrogel phantom as well as on ex vivo animal tissues. A 3D

blood vessel structure (a 10 mm vessel separates to two smaller branches with the width of 7 mm and 3.5 mm) was designed in SOLIDWORKS 2021 (Dassault Systemes, France), and printed using a 3D printer (Form 3L with the material Elastic 50A, Formlabs Inc., USA). The structure was then molded with 3.4 wt% gelatin hydrogel (gelatin from porcine skin, G1890, Sigma-Aldrich, Germany) mimicking a soft tissue (porcine brain) [19]. A fresh calf liver was obtained from a local butcher for ex vivo experiment. It was stored at 4 °C and used for experiments within 24 h after sacrifice of the animal. The motion of the robot was recorded by a Canon EOS RP camera with a 60 mm lens (Canon, Japan) and a USB camera (acA2440-75ucMED, Basler, Germany) with a lens 18-55 mm (Canon). The video was analyzed using ImageJ (1.53t, National Institutes of Health, USA).

3. Results

3.1. Characterization of magnetic field for actuation

Experimental- and simulation results of the magnetic flux density and the magnetic gradient are shown in Fig. 3. The sideview of the actuation system in XY plane is illustrated in Fig. 3a. M1 and M2 generate a magnetic field in the direction of +X and M3 generates a magnetic field that rotates around the X-axis. The center point of the two stationary magnets is defined as (0, 0, 0). At a vertical distance of 20 mm (i.e. the position (0, 20, 0), the robot achieves the fastest velocity while maintaining its walking stability (see Fig. 5 for detailed motion characterization). The working volume for a homogeneous magnetic field is defined by the angle difference of ± 10° from the angle at the measurement position. A working volume of 35 × 40 × 35 mm3 for the robot is achieved with the current set-up.

The magnetic flux densities in X-, Y- and Z-axes at the measurement position are presented in Fig. 3b. The magnet M3 rotates α from -75° to 75°, and the experimental and simulation results match perfectly at all angles of the M3. The simulation is performed in a specific range as the robot's walking motion angle of α is limited to ± 72° for a stable walking motion of the robot. Fig. 3c shows the magnetic flux density and the field gradient from the measurement point along the Y-axis at α = 0°. As the measurement point moves away from the rotating magnet, the magnetic flux density and magnitude of the magnetic gradient decrease. Both magnetic flux density and magnetic gradient show good agreement between measurements and simulation. The gradient is important to consider as it keeps the anchoring foot in firm contact with the surface, especially during the vertical climbing case (see Fig. 6b for details).

The numerical simulation results of the magnetic flux density in the YZ cross-section are shown in Fig. 3d. M1 and M3 are marked to show the position, and clockwise rotation of α is defined as positive. Initial pose of the robot is defined when α = 0°. The zero-degree of α is defined when the north pole of the magnet M3 pointing upwards along the Y-axis. The millirobot achieves the walking motion by rotating the magnet M3 with rotating angle α. The simulation results of the magnetic field vectors **B** and the corresponding robot's poses are shown in Fig. 3e.

3.2. Walking motion

The image sequence in Fig. 4 shows the robot's movement for a half cycle. The anchoring of the front feet is identical to the schematics as shown in Fig. 2a. Specifically, the front right (FR) foot is anchored at the start to facilitate a right turn. Then the front left (FL) foot touches the ground and anchors upon rotating back to α = 0°, enabling the next left turn. Due to the rotational motion along with anchoring mechanism, a displacement of the robot is observed every half cycle of actuation. The initial and last position are marked by orange and blue dashed line, respectively. The displacement is clearly visualized in the SI Video S1.

### 3.3. Locomotion characterization of the millirobot

Figure 5 shows the characterization results of the millirobot's locomotion under different conditions, by varying the oscillating angle, the pitch angle, the oscillating frequency, and the cargo weight, respectively.

The stride length of the millirobot was first measured while moving on the hydrogel phantom at different oscillating angles. Fig. 5a shows the experimental- and simulation-results. The averaged moving speeds range from 1.55 to 1.70 mm/cycle in the experiments, which match the simulation results very well with an average error of 10%. Experimental results exhibit a slightly lower stride length than the simulation because in the real case, the feet of the robot penetrate the soft material surface, which leads to a decrease of the real turning radius compared to the one used in the simulation. When α = 90°, the robot moves 1.8 % faster than α = 72°, however, the robot falls on the surface of hydrogel phantoms, when the angle α exceeds 72°. This is mainly due to surface tension of the fluid layer which was also observed when the robot moves on the biological tissues. To minimize the risk of being stuck at a location, the angle α is limited to 72° for all other experiments.

Fig. 5b shows the effect on the moving speed of the robot, when modifying the pitch angle θ from 39° to 66°. The speed increases almost linearly from 0.3 to 2.0 mm/s. The pitch angle is determined by the ratio between the magnitudes of $B_r$ and $B_h$ (Fig. 1b). While the positions of the magnets M1 and M2 are kept stationary, as the rotating magnet M3 approaches the robot, the magnetic field strength $B_r$ becomes larger, thus the pitch angle increases. The five pitch angles presented in Fig. 5b correspond to the measured positions in Y ranging from -20 to 20 mm in 10 mm steps. It was observed that the robot does not move when the pitch angle exceeds 70°, mainly due to the strong interaction between the rotating magnet and the robot's body. Therefore, further experiments are performed using a pitch angle of 66°, which provides a stable and fast crawling motion.

Experimental results also show that the robot's crawling speed also increases linearly with the oscillating frequency, as shown in Fig. 5c. It is challenging to maintain a stable walking motion over 1.5 Hz and consequently, an oscillating frequency of 1.2 Hz was chosen to ensure a stable walking motion of the millirobot.

As shown in Fig. 5d, the weight of the cargo has a tremendous effect on the walking speed. When the cargo weight is added from 0 to ~100 mg, the moving speed decreases from 2.0 to 1.4 mm/s. The robot's maximal load capacity is ~100 mg, which is approximately four times of its own body weight.

### 3.4. The control of the millirobot

The magnetic actuation system shown in Fig. 1b offers the possibility to control the in-plane crawling motion of the robot with two degrees of freedom to realize complex trajectories. The overlaid image of the robot's complex trajectory control videos (Video S2 and S3, playback speed at three times) is shown in Fig. 6a and Fig. 6b. During one trajectory, the maximum oscillating angle of α was set as 72° and oscillating frequency was fixed at a constant of 1.2 Hz. Complex trajectories were achieved by manually rotating the magnet set-up including M1, M2 and M3 by the angle β, and the sample table was kept stationary. The rotation of the set-up induces the rotation of the magnetic field around the Y-axis, which steers the robot to align with the external field. It allows the control of the robot to reach any in-plane target as shown in Fig. 6a. The robot is capable of rotating at the same position to make sharp turns, thus it offers the possibility of precise motion control.

Fig. 6b illustrates the time sequence of the robot climbing a vertical wall by anchoring to the surface by the magnetic gradient force generated by the magnet M3. The robot can crawl at a speed of 0.9 mm/s, facilitated by a perpendicular magnetic gradient force to the surface of 0.24 mN for anchoring. The gradient force is ~ 3 times larger than the gravitational force on the millirobot. It was also observed that the robot can walk upside down on a flat surface. The controllable crawling against gravity makes it suitable for biomedical applications in complex-shaped biological lumen systems.

3.5. Cargo deployment

Fig. 7 demonstrates the deployment of the robot's cargo on a hydrogel phantom and a calf liver. The robot was actuated forward by the oscillating magnetic field and executed the deployment scheme by rotating the angle α to 180°, as shown in Fig. 2b, to release the drug. Two types of cargos were used for demonstrations, i.e. a cotton pad for medical dressing and a capillary with a sharp tip for liquid drug injection into the surface of a soft tissue. After applying the cotton pad for 20 s, the red stain is found on the surface of the phantom (Fig. 7a). When the robot reaches a target position (Fig. 7b), it turns its body upside down for delivery. The magnetic gradient (0.1 T/m) generates a pulling force of ~ 0.3 mN on the robot, which provides enough force for the tip of the capillary to penetrate into the liver tissue.

4. Discussion

In this study, we present a new approach for a millimeter-scale robot to locomote on slippery biological surfaces to deliver drug using a customized magnetic actuation system. The robot shows the capability to walk and climb on various surfaces and deliver drugs using two types of cargo. The experimental results of the magnetic field and the motion of the millirobot are in good agreement with theoretical simulation.

We develop a permanent magnetic set-up for the wireless actuation of millirobots at the scale relevant to the human body. The set-up achieves an area of 50 cm x 15 cm and generates a flux density of ~7.5 mT at the center point. The average shoulder widths are 41 cm and 37 cm for male and female, respectively [20], which fit into the current system. The set-up can be extended to even larger distances by using larger permanent magnets. In a future surgical procedure, a human patient could be in a prone position, where the shoulders fit in between the two static magnets, and the anterior side is close to the rotating magnet. It shows the potential to use a similar set-up to actuate millirobots in the biological lumens in internal organs, such as the gastrointestinal tract and the biliary duct in the liver.

The walking motion of the millirobot is characterized under different conditions. Experimental results show that increasing the pitch angle or oscillating frequency can increase the robot's speed but may also lead to instability. The walking motion allows the millirobot to carry four times the body-weight, and also to crawl on a vertical surface of a hydrogel phantom. To allow the robot to climb on a vertical surface of real biological tissues, future work will be carried out to optimize the magnetic field, which enables the magnetic torque to be high enough to lift the robot's foot and the magnetic gradient force to be high enough to attract the robot to anchor to the vertical surface. By balancing these factors, we can design a millirobot that is optimized for a specific application.

The experimental result of the stride length during full fits the model of the robot's dynamics with one foot anchoring firmly to the surface. This result clearly shows that there is no slip during the 3D rotation of the millirobot, owing to the anchoring mechanism of the feet. One of the key strengths of our approach is the ability of the millirobot to move on a slippery biological surface in a controlled manner, which is in general a challenge for traditional small-scale robotic systems [21]. It is reported that the frictional coefficient of wet hydrogel is 95 times smaller than dry surfaces of a paper [22], which poses extra challenge for crawling robots based o n the friction to move on wet biological surfaces. Here, our millirobot utilizes sharp apices to penetrate soft tissues, which offers strong anchoring point for the robot's forward motion. Experimental results show that the anchoring is strong enough to hold its position on both hydrogel phantoms and calf liver tissues.

The anchoring mechanism helps the robot to hold its position on a slippery surface with the help of the magnetic gradient. It is observed from the experiment that only a robot with sharp tip achieve locomotion on a slipper surface not like the tumbling robots which generally require a rough surface such as poly(methyl methacrylate) [7] or dry paper [23]. The anchoring mechanism is based on the penetration of the sharp metallic tips into the soft tissue, which is protected by a wet mucus layer. In the experiments, permanent grooves on the order of 100 micrometers deep were observed on the surface layer. However, typical mucus layers, for example in the stomach, are around 1 mm thick [24], indicating the potential harmlessness. Future studies need to be carried out on real animal tissues to investigate the penetration depth of the feet to ensure safety in medical applications. Furthermore, to get closer to real biomedical applications, a consideration of biocompatibility is mandatory. Molybdenum, the material of robot's feet, has been reported as highly biocompatible, and other materials such as medical-grade stainless steel can also be applied to build the robot's feet. The magnetic materials NdFeB on the robots are not biocompatible, but they can be covered with biocompatible coatings [25].

The millirobot can readily switch between two types of motion, walking and cargo deployment. It can be controlled to navigate to an arbitrary in-plane target location, then deploy the carried drug, and it may move further to a second location for drug release at multiple locations. It is reported that small-scale robots can transport and release cargos [2,26]. However, to our knowledge, the cargos on a millirobot were normally released in the surrounding fluids but not directly into the soft tissues. Our deployment mechanism allows the injection directly into soft tissues which will be beneficial for many medical applications, especially for the retaining of drug at the target location.

## 5. Conclusions

In conclusion, our study shows the capabilities of a human-scale wireless magnetic actuation system for powering a millirobot to walk on slippery biological surfaces. The robot's walking and deployment motions are achieved through precise control of the magnetic field, enabling it to walk on various terrains and deliver cargo to desired positions. The versatility and adaptability of this magnetic actuation system makes it a promising platform for biomedical applications such as drug delivery and minimally-invasive medical procedures.

Supplementary Materials: The following supporting information can be downloaded at: www.mdpi.com/xxx/s1, Video S1: Walking motion, Video S2: Complex trajectory, Video S3: Vertical climbing, Video S4: Deployment on phantom, Video S5: Deployment on Liver.

Author Contributions: Conceptualization, M.J., X.T. and T.Q.; methodology, M.J. and T.Q.; analysis, M.J. and F.F.; writing—original draft preparation, M.J. and T.Q.; writing—review and editing, M.J., F.F. and T.Q.; supervision, T.Q.; funding acquisition, T.Q. All authors have read and agreed to the published version of the manuscript.

Funding: This work was partially funded by the Vector Foundation (Cyber Valley research group), the MWK-BW (Az. 33-7542.2-9-47.10/42/2) and the European Union (ERC, VIBEBOT, 101041975).

Data Availability Statement: All data needed to evaluate the conclusions in the paper are presented in the paper.

Acknowledgments: M. J. and T. Q. acknowledge the support by the Stuttgart Center for Simulation Science (SimTech). F. F. thanks the International Max Planck Research School for Intelligent Systems (IMPRS-IS) for the support.

Conflicts of Interest: The authors declare no conflict of interest.

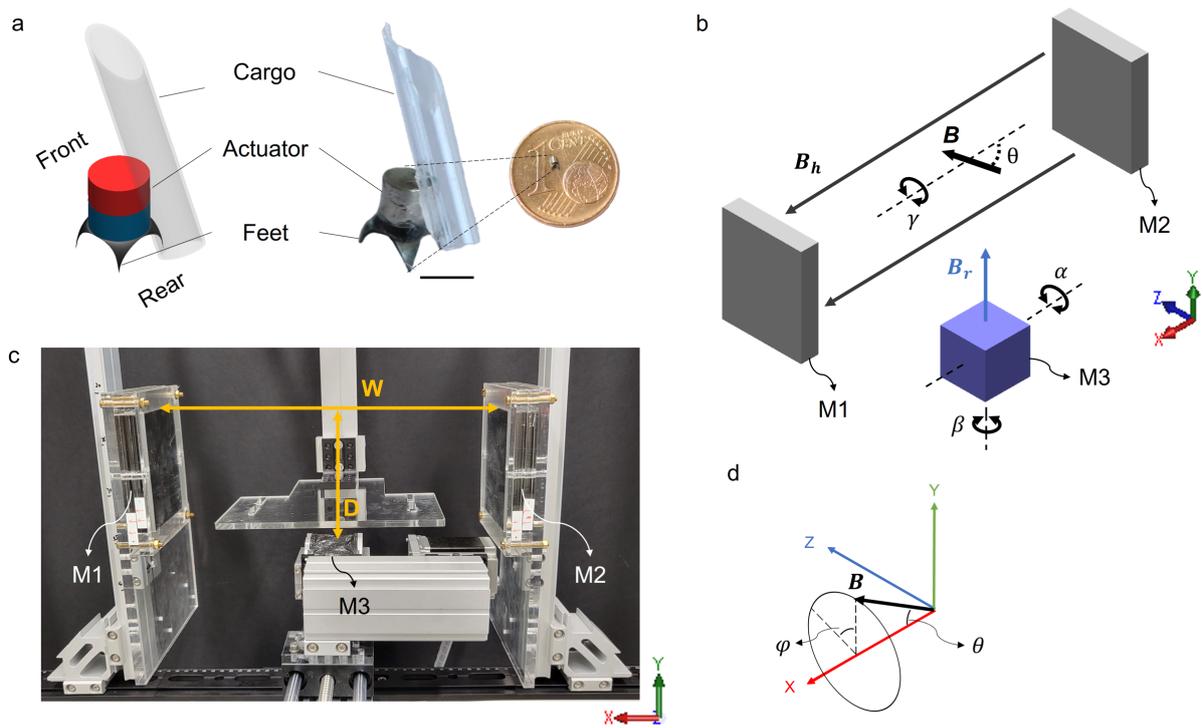

Figure 1. Design of the walking millirobot and the magnetic actuation set-up. (a) Schematic and image of the millirobot. The scale bar is 1 mm. (b) Schematic of the actuation system based on a pair of stationary magnets M1 and M2 and a rotating magnet M3. (c) Picture of the magnetic actuation system. (d) Diagram showing the superimposed magnetic field vector at the center of the working volume.

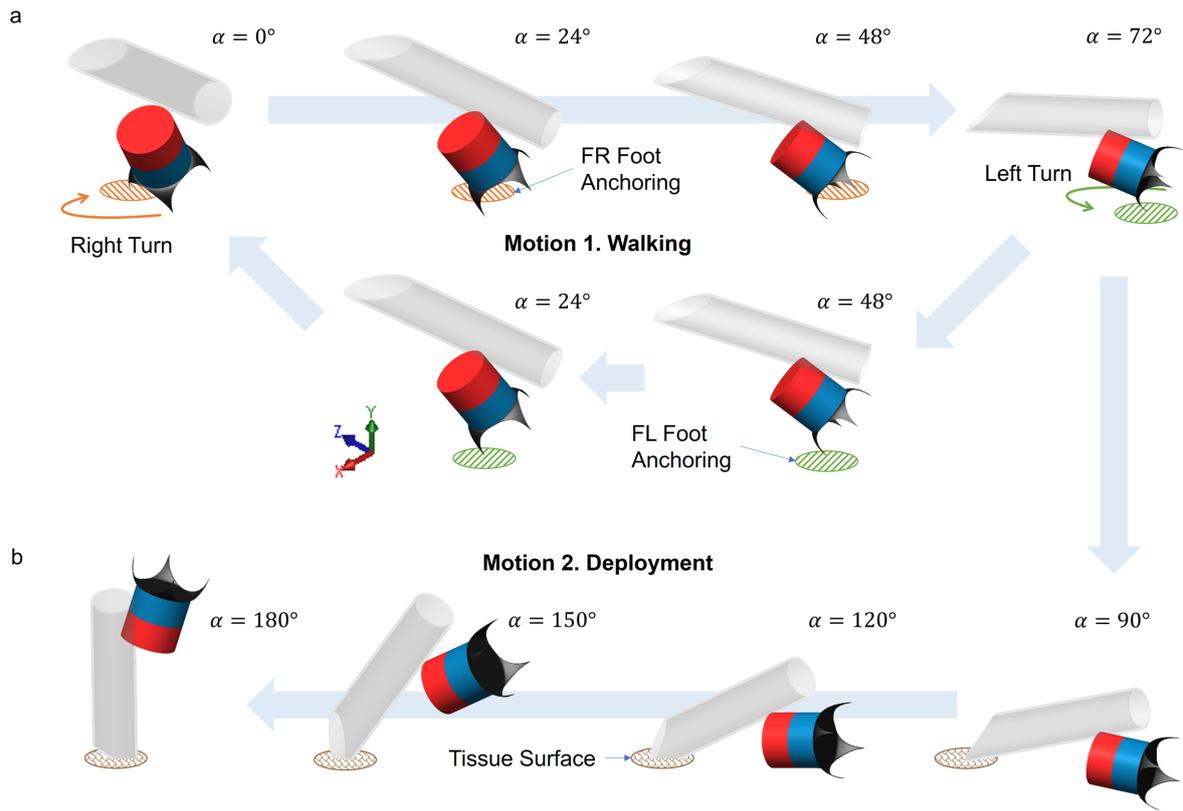

Figure 2. Motion schemes of the millirobot. (a) Motion 1 shows a half cycle of the gait of the robot to move forward. The oscillating angle α for walking is limited to maximally 72° to prevent the body from touching the ground and to stabilize the gait of the robot. (b) Motion 2 shows the cargo deployment scheme, e.g. drug injection into the tissue by the capillary. FL: front left, FR: front right.

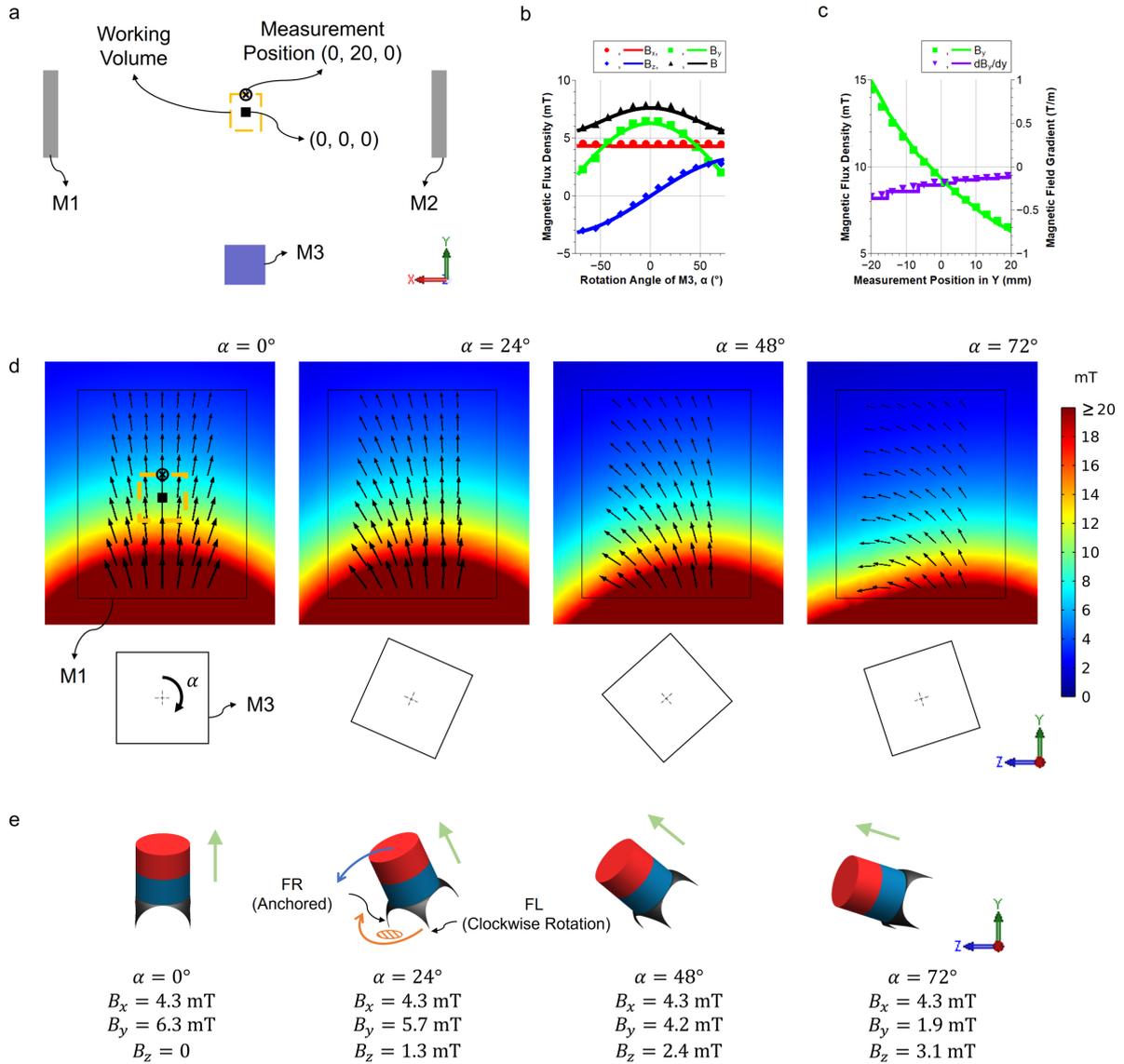

Figure 3. Experimental and simulation results of the magnetic actuating system. (a) Schematic of the system. The working volume (35 x 40 x 35 mm$^3$) and the measurement position (0, 20, 0) are marked in the yellow box and the black cross, respectively. (b) Magnetic flux density of the experimental (dots) and simulation results (Solid lines) at the measurement position (-75° ≤ α ≤ 75°). (c) Magnetic flux density and magnetic gradient along the Y-axis for an actuation angle of α = 0°. Solid lines and dots represent simulation and measured data, respectively. (d) Simulation result of the magnetic flux density at different angles of α. The north pole of M3 points upwards along with Y-axis when α = 0 °. The colormap indicates the magnitude of the magnetic flux density B. (e) Millirobot's poses and corresponding magnetic flux density at the measurement position. The green arrow indicates the direction of the external magnetic field.

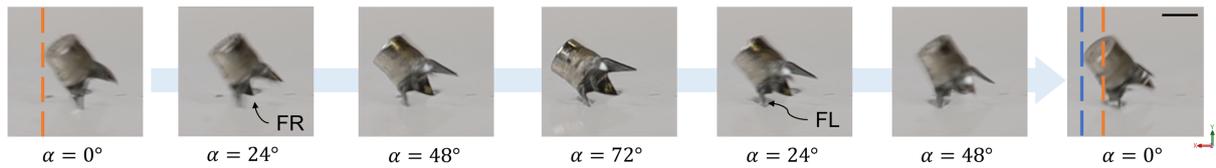

Figure 4. Snapshots of the walking motion during a half cycle. The yellow dashed line labels the start position, and the blue dashed line labels the end position after a half cycle. The scale bar is 1 mm.

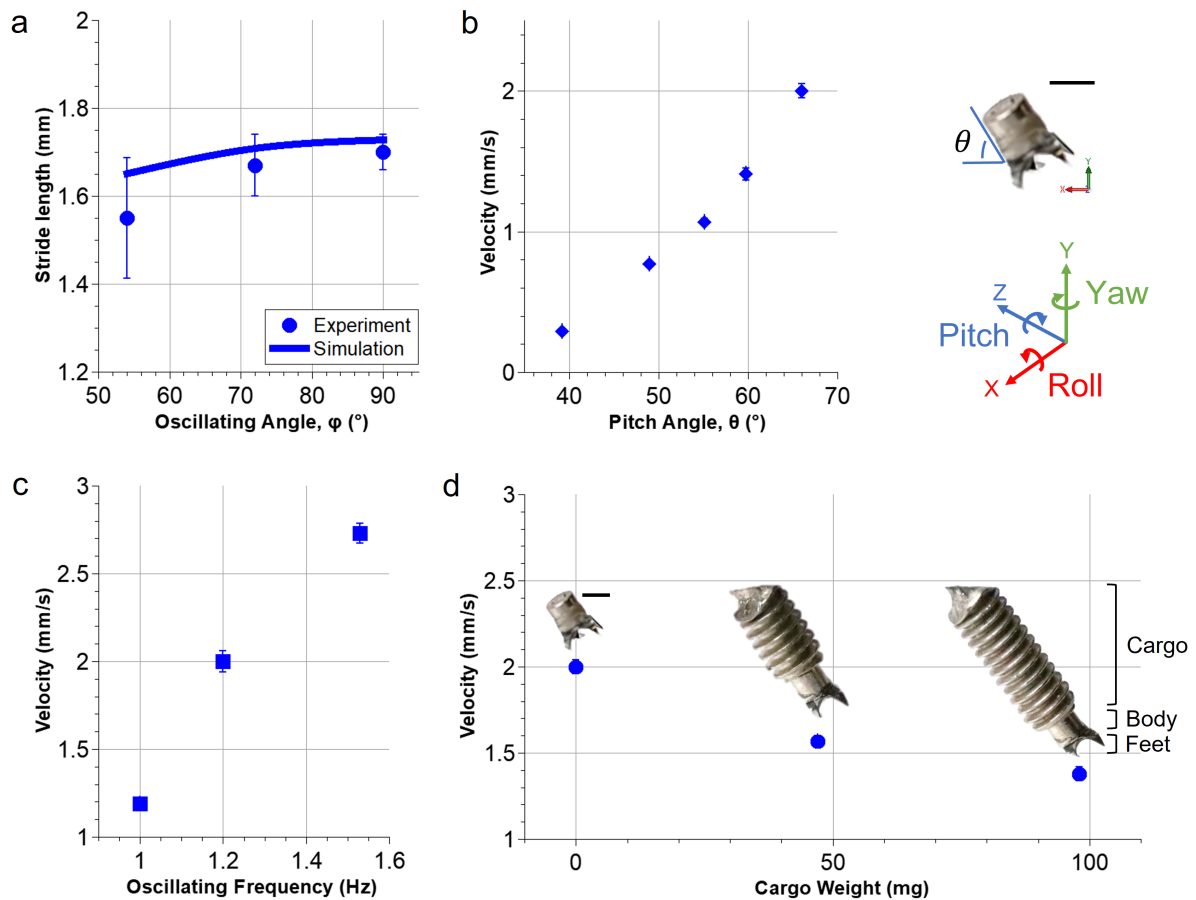

Figure 5. Characterization of the locomotion of the millirobot. (a) The stride length at different oscillating angles. (b-d) The moving velocity at a constant oscillating angle of 72° but varying the pitch angle θ (b); the oscillating frequency (c); and the cargo weight (d). All scale bars are 1 mm.

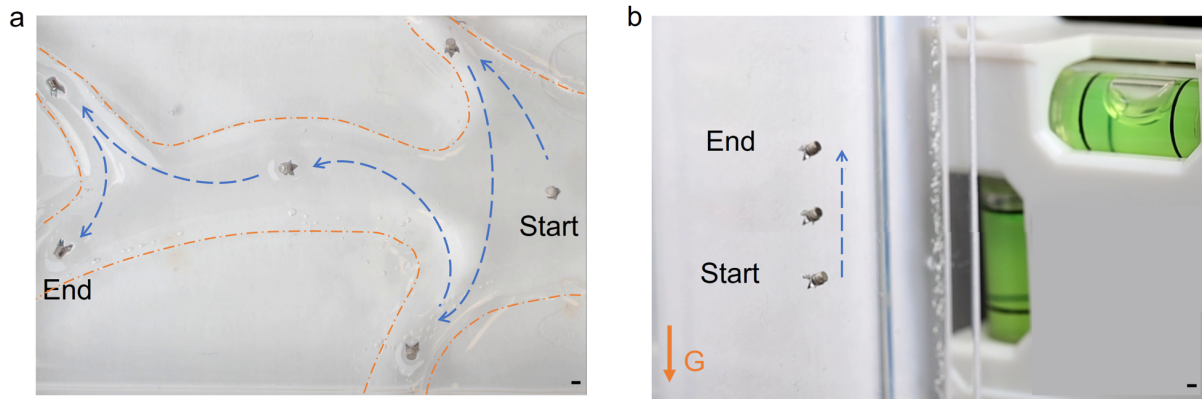

Figure 6. Control of the millirobot to achieve complex trajectories on hydrogel phantoms. (a) The robot follows a complex trajectory. Blue dashed lines depict the moving trajectory, and yellow dashed line highlights the boundary of the phantom. (b) The robot climbs up a vertical wall. The arrow G indicates the direction of gravity. All scale bars are 1 mm.

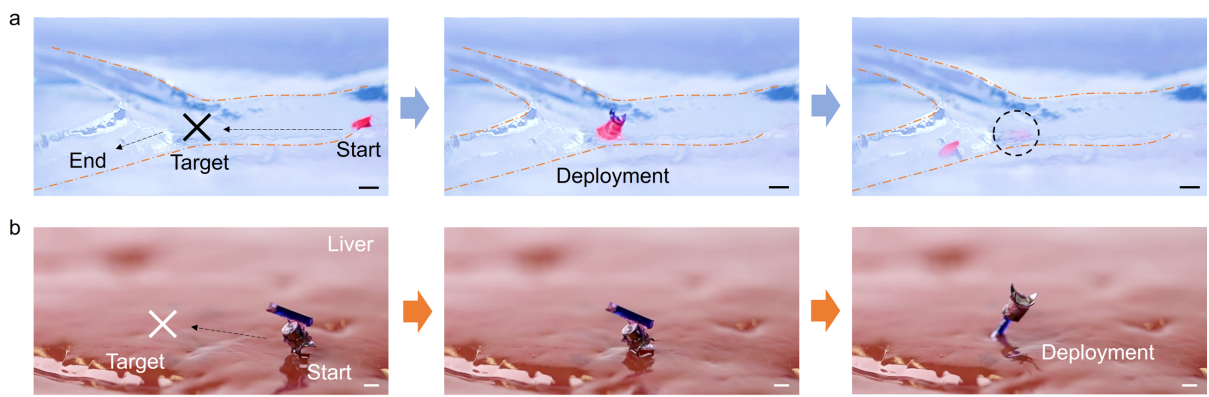

Figure 7. Targeted drug delivery by a millirobot. (a) Application of a drug using a cotton pad on a gel phantom (Video S4, playback speed at three times). (b) Locomotion of the robot on a liver tissue to a target position and injection of a drug into the liver tissue by capillary (Video S5). The target location is labeled by a cross and the dashed circle highlights the applied drug on the surface. All scale bars are 1 mm.